\documentclass[journal]{IEEEtran}

\usepackage{xcolor,soul,framed} 

\colorlet{shadecolor}{yellow}
\usepackage[pdftex]{graphicx}
\graphicspath{{../pdf/}{../jpeg/}}
\DeclareGraphicsExtensions{.pdf,.jpeg,.png}

\usepackage[cmex10]{amsmath}
\usepackage{array}
\usepackage{mdwmath}
\usepackage{mdwtab}
\usepackage{eqparbox}
\usepackage{url}
\usepackage{color}
\usepackage{cite}
\usepackage{amssymb,amsfonts}
\usepackage{multirow}
\usepackage{subcaption} 
\usepackage{float} 
\usepackage{algorithm}
\usepackage{algpseudocode}
\usepackage{color}
\usepackage{booktabs}
\usepackage{lineno}
\usepackage{hyperref}
\begin{document}
\bstctlcite{IEEEexample:BSTcontrol}
\title{Automated Dilated Spatio-Temporal Synchronous Graph Modeling for Traffic Prediction}
\author{Guangyin Jin*,
        Fuxian Li*,
        Jinlei Zhang,
        Mudan Wang and
        Jincai Huang
\IEEEcompsocitemizethanks{\IEEEcompsocthanksitem F.~Li,  M.~Wang are with Beijing National Research Center for Information Science and Technology (BNRist), Department of Electronic Engineering, Tsinghua University, Beijing 100084, China.\protect\\
E-mail: lifx19@mails.tsinghua.edu.cn, wmd20@mails.tsinghua.edu.cn
\protect\\
\IEEEcompsocthanksitem G.~Jin and J.~Huang are with College of Systems Engineering, National University of Defense Technology, Changsha, China.\protect\\
E-mail: jinguangyin18@nudt.edu.cn, huangjincai@nudt.edu.cn\protect\\
\IEEEcompsocthanksitem J.~Zhang is with State Key Laboratory of Rail Traffic Control and Safety, Beijing Jiaotong University, Beijing 100044, China.\protect\\
E-mail: zhangjinlei@bjtu.edu.cn\protect\\
\IEEEcompsocthanksitem $^{*}$Both authors contributed equally to this research.
}}


\maketitle

\begin{abstract}
Accurate traffic prediction is a challenging task in intelligent 
transportation systems because of the complex spatio-temporal dependencies in transportation networks. Many existing works utilize sophisticated temporal modeling approaches to incorporate with graph convolution networks (GCNs) for capturing short-term and long-term spatio-temporal dependencies. However, these separated modules with complicated designs could restrict effectiveness and efficiency of spatio-temporal representation learning.  Furthermore, most previous works adopt the fixed graph construction methods to characterize the global spatio-temporal relations, which limits the learning capability of the model for different time periods and even different data scenarios. To overcome these limitations, we propose an automated dilated spatio-temporal synchronous graph network, named Auto-DSTSGN for traffic prediction. Specifically, we design an automated dilated spatio-temporal synchronous graph (Auto-DSTSG) module to capture the short-term and long-term spatio-temporal correlations by stacking deeper layers with dilation factors in an increasing order. Further, we propose a graph structure search approach to automatically construct the spatio-temporal synchronous graph that can adapt to different data scenarios. Extensive experiments on four real-world datasets demonstrate that our model can achieve about $10\%$ improvements compared with the state-of-art methods. Source codes are available at  https://github.com/jinguangyin/Auto-DSTSGN.
\end{abstract}


%
\IEEEpeerreviewmaketitle


\section{Introduction}\label{sec:intro}
\IEEEPARstart{T}{raffic} prediction plays a basic but crucial role in intelligent transportation systems, which has been widely deployed in some online services such as navigation and ride-hailing. 
Effective spatio-temporal modeling is key to obtain more precise predictions. In recent years, most works utilize spatial modules including convolutional neural networks (CNNs) and graph neural networks (GNNs), and temporal modules such as recurrent neural networks (RNNs), temporal convolution networks (TCNs) for spatio-temporal modeling~\cite{stgat,stgcn,stgnn,stmetanet,dcrnn,gwn,jin2021spatio,astgcn,liu2020dynamic,liu2020short,liao2022taxi,ikram2022prediction,jin2022stgnn,jin2022adaptive,james2022graph,jin2021gsen,jin2020ufsp,liu2020physical,jin2020fire}, but the complex correlations in spatio-temporal scale are still difficult to learn. 

\begin{figure}[t]
\centering
\vspace{-2mm}
\includegraphics[width=0.43 \textwidth]{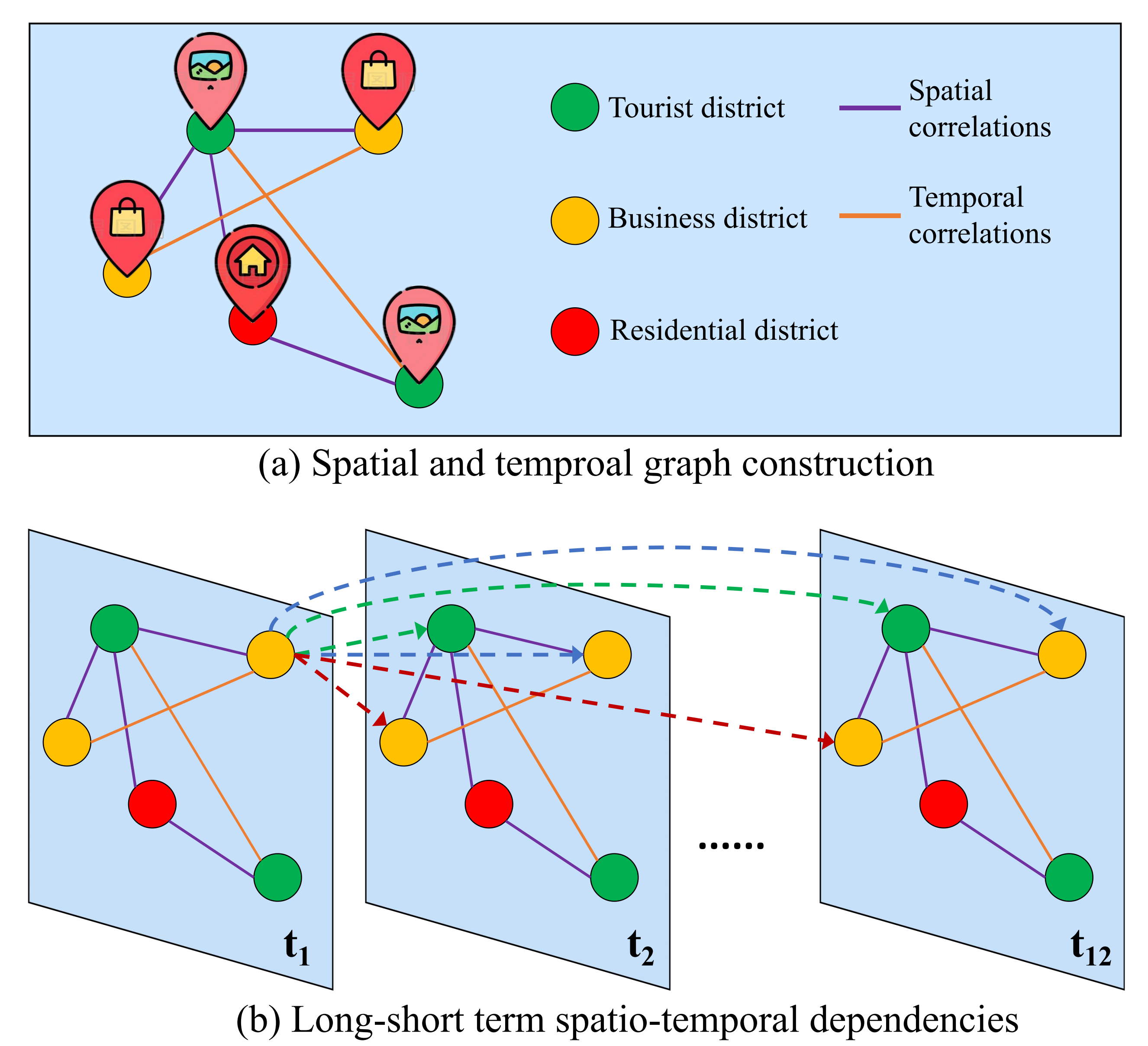}
\caption{Example of spatial and temporal graph construction and spatio-temporal dependencies in a network. The blue dotted arrows, green dotted arrows and red dotted arrows respectively represents the cross-time self-connection, cross-time geo-spatial relations and cross-time pattern similarities.}
\label{fig:intro} 
\vspace{-3mm}
\end{figure}

First, both geo-spatial relations and pattern similarities exist in traffic network at the same time. As shown in Figure~\ref{fig:intro}(a), We can respectively construct the spatial graph and temporal graph to characterize the proximity (e.g., tourist district and business district) and similar patterns (e.g., two different tourist districts) between different nodes. Second, the two different dependencies not only exist in the same time interval but also have influences across different time steps. The correlations across different time steps can be short-term (e.g., time step $t_1$ to $t_2$) or long-term (e.g., time step $t_1$ to $t_{12}$), as shown in Figure~\ref{fig:intro}(b). Although many graph-based deep learning models have been demonstrated effective in traffic prediction, there are still at least two limitations. 

(a) \textbf{Capturing complex long-short term spatio-temporal dependencies.} Most works employ GNNs and some temporal learning components to respectively capture the spatial and temporal correlations~\cite{ye2020build}. For temporal modeling, RNN is a classical method but suffers from gradient vanishing or explosion for learning long-range sequences~\cite{servan1988learning}.
Some superior variants such as LSTM~\cite{lstm} and GRU~\cite{Cho2014LearningPR} alleviate the gradient problem to some extent, but still suffer from low computational efficiency due to the recurrent structures.
Self-attention mechanism such as Transformer~\cite{transformer,stgnn} is designed to capture long-term dependencies but it is still time-consuming during training and inference phase. TCN considers both performance and efficiency but it is hard to capture long-term dependencies by fixed kernel sizes~\cite{stgcn}. To address this problem, Graph WaveNet~\cite{gwn} integrates GCN and dilated TCN to capture long-short term dynamics by stacking layers with the dilation factors in an increasing order. However, the framework with separate modules is still hard to capture more complex dependencies from spatio-temporal scale, as shown in Figure~\ref{fig:intro}(b). To enhance the learning capability for spatio-temporal dependencies, STSGCN~\cite{stsgcn} and STFGNN~\cite{stfgnn} construct the spatio-temporal synchronous graphs (STSGs). However, the fixed receptive field of them limits their capability for learning both short-term and long-term spatio-temporal dependencies.

(b) \textbf{Flexible spatio-temporal graph construction.} In most previous works, spatio-temporal graph is artificially designed and fixed for different time periods~\cite{stgcn,stsgcn,dcrnn,stfgnn,jin2020urban,jin2021hierarchical,jin2020deep}. Meanwhile, the method of graph construction cannot be adjusted even on different datasets. Hence, this manner can hardly characterize diverse spatio-temporal relations for different time periods and different datasets. The learnable mask~\cite{stsgcn,stfgnn} and adaptive graph~\cite{gwn,mtgnn,staggn,jin2022adaptive,li2021dynamic} are introduced to overcome the limitation of fixed spatio-temporal graph but they still have many disadvantages. On one hand, the modeling of them is global, thus they fail to characterize spatio-temporal relations for different time periods. On the other hand, they cannot be associated with the characteristics of the data itself, hence they have weak interpretability. 

To address the above problems, we propose a novel framework for traffic prediction, called Automated Dilated Spatio-Temporal Synchronous Graph Network (Auto-DSTSGN). To be specific, we design the dilated spatio-temporal synchronous graph framework to flexibly capture the short-term and long-term spatio-temporal complex dependencies. Further, a graph structure search operation is proposed to construct the flexible and diverse STSGs automatically according to different data input in different time periods. Our main contributions in this paper are summarized as follows:
\begin{itemize}
\item We design a dilated spatio-temporal synchronous graph framework to capture spatio-temporal correlations efficiently, whose receptive field can become larger by stacking deeper layers with the dilation factors in an increasing order. This framework can capture both short-term and long-term dependencies with relatively low computation burden and GPU occupancy. 
\item We propose the graph structure search operation based on DARTS framework. As far as we know, it is the first attempt to search adjacency matrices rather than neural architectures by auto machine learning methods.
\item We conduct extensive experiments on four public traffic datasets. The experimental results demonstrates that our model can obtain at least 4.9\%$\sim$10.3\% improvements compared with the state-of-art baselines. 
\end{itemize}


\section{Related Work}
\subsection{Traffic Prediction}

In recent years, spatio-temporal graph modeling has become a mainstream method for traffic prediction. Most of these works combine spatial graph convolution networks (GCNs) with some temporal encoder to capture the complex spatio-temporal dependencies. STGCN~\cite{stgcn} first integrate TCNs and GCNs for spatio-temporal modeling. Based on this, ASTGCN~\cite{astgcn} adopts attention mechanism to enhance the representation learning capability of GCNs and TCNs, Graph WaveNet~\cite{gwn} involves the adaptive graph to incorporate with dilated temporal convolution networks. RNN is a 
widely used model for sequence leaning, whose variant gated recurrent unit (GRU) is utilized in DCRNN~\cite{dcrnn} and T-GCN~\cite{tgcn} to capture temporal correlations of hidden representations from GCNs. In addition, self-attention mechanism is powerful tool not only for spatial dynamic learning, but also for capturing temporal dependencies~\cite{transformer}. Both STGNN~\cite{stgnn} and GMAN~\cite{gman} adopt self-attention mechanism in spatial GCNs and temporal dependencies learning. In some most recent works, some novel frameworks are introduced in this field. STSGCN~\cite{stsgcn} first proposes the framework of spatio-temporal synchronous modeling. Based on this, STFGNN~\cite{stfgnn} presents an informative fusion graph and parallel TCNs for further improving spatio-temporal dependencies learning. AutoSTG~\cite{autostg} first combines neural architecture search approach with spatio-temporal graph framework to improve the adaptability for different data. However, these existing works are not only difficult to flexibly capture the long-short term complex spatio-temporal dependencies, but also hard to characterize the spatio-temporal relations in different periods. Different from them, our model can balance short-term and long-term spatio-temporal correlations by dilation mechanism, and the graph search operation in our model can construct diverse spatio-temporal relations automatically  for different datasets.

\subsection{Automated Machine Learning}

Automated Machine Learning (AutoML) aims to obtain appropriate features or models for various downstream tasks~\cite{automl}. This field can be roughly divided into two main categories: automated feature engineering and automated model design, while automated feature engineering aims to synthesize or select informative features for model training~\cite{autof,autof1,autotime}. Automated model design aims to select appropriate models or construct reasonable model architecture. Neural architecture search (NAS) is one of the most important direction in automated model design, which is widely used in deep learning. There are three mainstream types of methods in NAS, reinforcement learning-based~\cite{liu2018progressive,pham2018efficient,zoph2016neural}, evolutionary learning-based~\cite{real2019regularized,sun2020automatically,xie2017genetic} and gradient-based~\cite{liu2018darts,chen2019progressive,li2020autost} respectively. Among these three types, gradient-based methods are relatively more efficient. Since efficiency is important in traffic prediction, we adopt gradient-based framework DARTS~\cite{liu2018darts} in this paper. Different from these previous works, we employ DARTS to search graph structure rather than neural architecture in this paper.

\section {Preliminary}

\subsection{Problem Definition}
Given the traffic sensor set with $N$ nodes $V(\vert V\vert = N)$, the sensor network can be defined as a graph $\mathcal{G} = (V, E, A)$. $E$ denotes the set of edges, whose relations between different nodes is characterize by the adjacency matrix $A$.
The graph signal at time step $t$ contains $d$-dimensional original traffic features (e.g., the speed, volume), which is defined as ${X}_{\mathcal{G}}^{(t)} \in \mathbb{R}^{N \times d}$. The aim of traffic prediction task in sensor network is to learn a non-linear function $f(\cdot)$ from historical $T$-step graph signals for forecasting next $T^{'}$-step graph signals. The mathematical form is defined as follows:
\begin{equation}
[\mathbf{X}_{\mathcal{G}}^{(t-T+1)}, \cdots, \mathbf{X}_{\mathcal{G}}^{t}] \xrightarrow[]{f(\cdot)} [ \mathbf{X}^{t+1}_{\mathcal{G}}, \cdots, \mathbf{X}^{t+T^{'}}_{\mathcal{G}}]\label{d1}.
\end{equation} 

\begin{figure}[htb]
\centering
\vspace{-3mm}
\includegraphics[width=0.43 \textwidth]{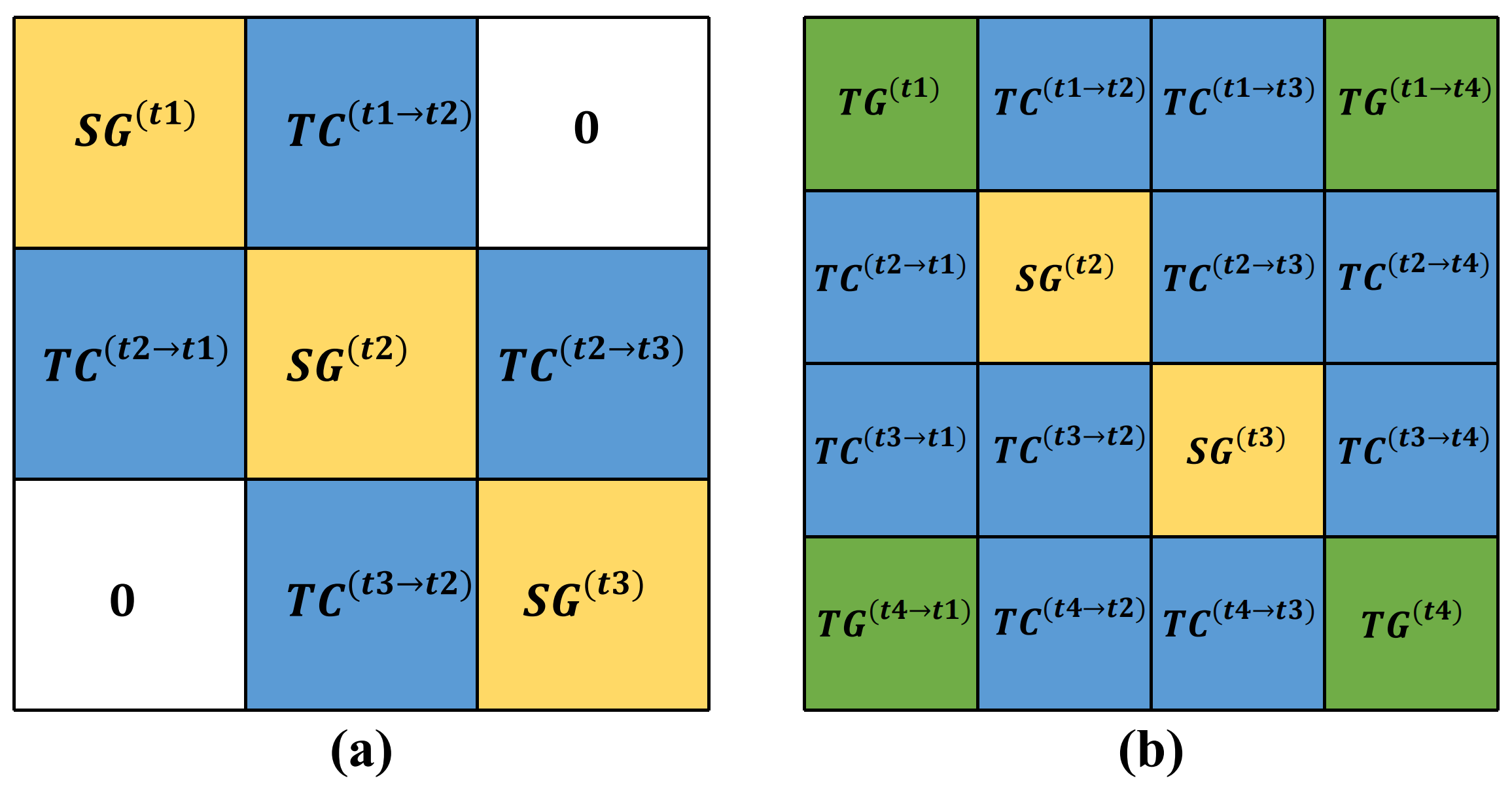}
\caption{Spatio-temporal synchronous graph construction in previous works. (a) is the adjacency matrix in STSGCN and (b) is the adjacency matrix in STFGNN. $SG^{t_i}$ and $TG^{t_i}$ respectively denote the spatial graph and temporal graph at time step $i$. $TC^{t_i\to t_j}$ describes the self-connectivity of nodes at the time step $i$ and $j$, while $TG^{t_i\to t_j}$ denotes the temporal-pattern similarities among nodes at the time step $i$ and $j$.}
\label{fig:sts_model} 
\vspace{-3mm}
\end{figure}

\begin{figure*}[h]
\centering
\vspace{-3mm}
\includegraphics[width=0.85 \textwidth]{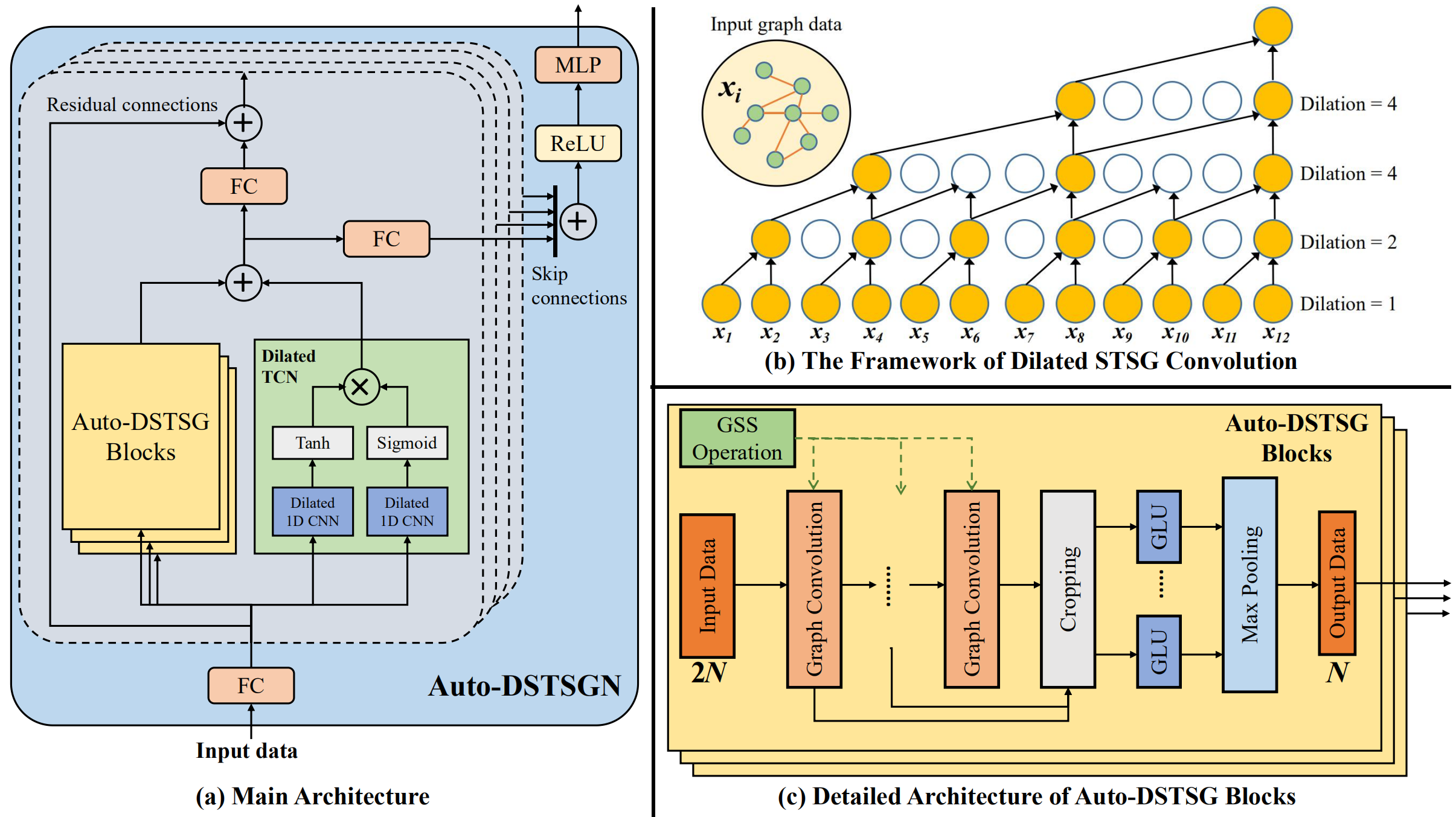}
\caption{Detailed framework of Auto-DSTSGN. (a) is the main architecture of Auto-DSTSGN. The Automated Dilated Spatio-Temporal Synchronous Graph Convolution (Auto-DSTSG) Module and the Dilated Temporal Convolution (DTC) Module filter the input of each layer in parallel to obtain comprehensive spatio-temporal modeling.  (b) is a unified framework of dilated convolution with kernel size 2. With a dilation factor d, the input data is picked every d steps, which is marked by yellow. The receptive field becomes larger with dilation factor increasing. (c) is the detailed  architecture of Auto-DSTSG blocks. In each block, the Graph Structure Search (GSS) operation is an auto machine learning-based operator to construct diverse Spatio-temporal Synchronous Graph (STSGs) according to different input data. }
\label{fig:overview} 
\vspace{-3mm}
\end{figure*}

\subsection{Meta Graph Construction}
In this paper, the graph that characterizes the spatio-temporal relations in one time step is called as the meta graph, which is the basic unit of STSGs. To take both geographical proximity and pattern similarity into account, we introduce two types of meta graph: spatial graph $A_{SG} \in \mathbb{R}^{N \times N}$ and temporal graph $A_{TG} \in \mathbb{R}^{N \times N}$. 
The adjacency matrix of spatial graph can be formulated as:
\begin{equation}
  A^{ij}_{SG}=\left\{\begin{aligned}
  1, &\ if \ v_{i} \ connects \ to \ v_{j}  \\
  0, &\ \text{otherwise}
  \end{aligned},
  \right.
\end{equation}
Dynamic Time Warping (DTW) algorithm is adopted to calculate the similarity of two time series~\cite{dtw}, which can characterize the pattern similarity between different nodes. 
For example, given two time series $X=(x_1, x_2, \cdots, x_m)$ and $Y=(y_1, y_2, \cdots, y_n)$, DTW is a dynamic programming algorithm defined as:
\begin{small}\begin{equation}
  D(i, j) = \vert x_i-y_j\vert + \min \left( D(i-1,j),D(i, j-1),D(i-1,j-1)\right),
\end{equation}\end{small}
where $D(i,j)$ denotes the shortest distance between sub-sequence $X_s=(x_1, x_2, \cdots, x_i)$ and $Y_s=(y_1, y_2, \cdots, y_j)$. As a result, we can obtain $DTW(X,Y)=D(m,n)$ as the final distance between $X$ and $Y$. This method not only does not need to limit the length of the input sequences, but also better reveals the similarity of two time series compared with the Euclidean distance.
Thus, we can define the adjacency matrix of temporal graph through the DTW distance as following:
\begin{equation}
  A^{ij}_{TG}=\left\{\begin{aligned}
  1, &\ DTW(X^i, X^j) < \epsilon \\
  0, &\ \text{otherwise}
  \end{aligned},
  \right.
\end{equation}
where $X^i$ and $X^j$ are speed data series attached to node $i$ and node $j$ respectively. $\epsilon$ is a threshold to control the sparsity of $A_{TG}$, the setting of which is the same as that in~\cite{stfgnn}.

\subsection{Spatio-Temporal Synchronous Graph Modeling}
Spatio-temporal synchronous graph (STSG) is a special architecture to establish the unified spatio-temporal correlations by graph structures, which is first proposed in~\cite{stsgcn}. Given a spatial graph with $N$ nodes, the adjacency matrix of STSG in~\cite{stsgcn} is designed as a $3N\times 3N$ expanded matrix, as shown in Fig.~\ref{fig:sts_model}(a). In~\cite{stfgnn}, the STSG is extended to $4N\times 4N$ and improved by involving more informative graphs such as temporal graph, which is shown in Fig.~\ref{fig:sts_model}(b). Although this modeling approach can characterize more complex spatio-temporal relations, the larger expanded STSG could bring the problem of high computational overhead.

\section{Methodology}
The overview of our model is illustrated in Figure~\ref{fig:overview}.
To enhance the capability of representation learning, we employ a fully connected layer at the top of our model to transform the input graph signal into high-dimensional space.
Then we design multiple stacked layers for extracting spatio-temporal structural information.
At each layer, we design Automated Dilated Spatio-Temporal Synchronous Graph Convolution (Auto-DSTSG) Module with multiple parallel blocks for modeling complex spatio-temporal dependencies. 
To be specific, Auto-DSTSG Module in each layer can expand receptive field to capture both short-term and long-term spatio-temporal complex dependencies by the increased dilation factors, which addresses the first limitation in Sec.~\ref{sec:intro}. 
In each parallel Auto-DSTSG block, we also propose the Graph Structure Search (GSS) Operation to automatically construct adjacency matrix of spatio-temporal synchronous graph (STSG), which can characterize the flexible and diverse spatio-temporal relations in different time periods. This deals with the second limitation in Sec.~\ref{sec:intro}. Then the graph convolution layers are adopted to capture the spatio-temporal correlations based on the constructed STSGs. There are also many supporting operators in each block: cropping, gated linear units (GLUs) and max pooling. The cropping operation is to ensure dimensional consistency with the output. The gated linear units are used to increase the nonlinearity of graph convolutional features. And max pooling is to preserve the most distinctive features. Their details are shown in the subsection~\ref{sec:adsts}.
In addition, we also design Dilated Temporal Convolution Module to enhance the global temporal correlations, whose output is aggregated with the output from Auto-DSTSG Module in each layer.
We further adopt the residual connection and skip connection in each layer of Auto-DSTSGN. The hidden information of each layer flows to the next layer after passing through residual connections. On the other hand, the hidden information of each layer is output through skip connections. At the top of the framework, the outputs from the skip connections of each layer are summed up as the input to the MLP to obtain the predictions.

In the following subsections, we introduce Automated Dilated Spatio-Temporal Synchronous Graph Convolution Module, Graph Structure Search Operation, Dilated Temporal Convolution Module and some other components in details. 
In addition, we also show a brief overview of the optimization algorithm of graph structure search.
To facilitate understanding of the following expressions, we list the definition of some important notations and operators throughout the overall methodology in Table~\ref{tab:definition}.
\begin{table}[h]
\centering
\caption{The definition of some important notations and operators in our methodology.}
\vspace{-3mm}
	\scalebox{1.1}{
	\begin{tabular}{cc}
		\hline
		Symbols  & Definition \\ \hline
		$\mathcal{F}_{G}(\cdot)$   & Graph structure search operation\\
		$A_{ST}$    & Final adjacency matrices of STSGs  \\
		$\mathcal{M}_{A}$    &  Mixed adjacency matrices of STSGs  \\
		$\mathcal{M}_{1}$    & Mixed main-diagonal matrices of STSGs  \\ 
		$\mathcal{M}_{2}$    & Mixed sub-diagonal matrices of STSGs \\
		$\mathcal{C}(\cdot)$    & Cropping operation for STSGs\\ \hline
	\end{tabular}}
	\label{tab:definition}
\vspace{-3mm}
\end{table}

\subsection{Automated Dilated Spatio-Temporal Synchronous Graph Convolution Module}~\label{sec:adsts}
Recall that a normal spatial GCN layer~\cite{kipf2017semi} has the form:
\begin{equation}
    \mathbf{Z} = \sigma(\hat{\mathbf{A}}\cdot \mathbf{X}\cdot \mathbf{\Theta}),
\label{eq:gcn}  
\end{equation}
where $\mathbf{X}\in R^{N\times D}$ and $\mathbf{Z} \in R^{N\times D'}$ respectively denote the input and output node embedding of the GCN layer. $\mathbf{\Theta}\in R^{D\times D'}$ denotes the shared weight for nodes' feature mapping and $\sigma(\cdot)$ denote the activation function. $\hat{\mathbf{A}}\in R^{N\times N}$ denotes the normalized adjacency matrix which represents the message passing between one-hop neighbors.

To capture the complex spatio-temporal correlations, the normal spatial GCN can be extended to the spatio-temporal scale. A special GCN-based framework called spatio-temporal synchronous graph convolution network is proposed~\cite{stsgcn}, but there are two main limitations of this modeling approach. 
The first one is that the fixed spatio-temporal receptive field limits the learning capability for long-term dependencies. 
To capture the longer-term spatio-temporal dependencies by this framework, the receptive field should become larger. Whenever the receptive field is expanded by one unit, the calculation amount of matrix multiplication will be expanded exponentially. 
The second limitation is that the adjacency matrix of STSG is shared for each spatio-temporal synchronous graph convolution module, which can not reflect the diverse spatio-temporal correlations in different periods. In addition, the deterministic adjacency matrix designed manually is hard to characterize the complex relations between different spatio-temporal nodes for different datasets.

To overcome these problems, we design the Auto-DSTSG module with multiple parallel blocks. 
The form of the Auto-DSTSG block is defined as:
\begin{align}
    \label{eq:dsts}
    &\mathbf{Z} = \sigma(\mathcal{C}(\mathcal{F}_{G}(\mathbf{X}(t, k, d))\cdot \mathbf{\Theta})),\\
    &\mathbf{X}(t, k, d) = [\mathbf{x}(t-d\times(k-1)),\dots,\mathbf{x}(t-d), \mathbf{x}(t)],
\end{align}
where $\mathbf{X}(t, k, d)\in R^{kN\times D}$ and $\mathbf{Z} \in R^{N\times D'}$ respectively denote the input and output representation. $k$ denotes the kernel size and $d$ denotes the dilation factor to control the skipping distance in the sequence input. $[\cdot]$ denotes the concentration operation. Compared with the normal GCN, the receptive field has been expended to $d\times k$ times when the range of spatio-temporal synchronous graph covers $k$ time steps. Here $\mathcal{F}_{G}(\cdot)$ represents the graph structure search operation, whose output is a mixed adjacency matrix $\mathcal{M}_{A} \in R^{kN\times kN}$ during searching phase. This operation can adjust the structure of STSGs according to different data, which can characterize complex spatio-temporal relations in different periods and even different datasets.
$\mathcal{C}(\cdot)$ denotes cropping operation to convert the range of data from $kN$ to $N$. The common method is to select the graph data of the middle or last time step as the final output, which is described in~\cite{stgnn}. When $\mathcal{F}_{G}(\cdot)$ is a fixed adjacency matrix with $k=1, d=1$ and $\mathcal{C}(\cdot)$ is an identity mapping, eq~(\ref{eq:dsts}) is equivalent to eq~(\ref{eq:gcn}). 


\subsubsection{Dilated Spatio-temporal Synchronous Graph Convolution Framework}
The framework of stacked dilated temporal convolution was first proposed in~\cite{yu2016multi}. To be specific, the long-term correlations can be captured effectively by stacking deeper layers with dilation factors in an increasing order. Inspired by this, we expand the framework to the case of spatio-temporal synchronous graph convolution. In this framework, since even small kernel size can be competent for long-term dependencies learning, we do not need large size STSGs to cover long-range spatio-temporal correlations. The computational burden and memory occupancy can also be reduced greatly by small kernel sizes.
Thus, the kernel size $k$ is fixed as 2 in our model. Suppose the input length of our model is 12 time steps, we can design a four-layer framework with a sequence of dilation factors $[1,2,4,4]$ to cover the whole receptive field of the 12 time steps, as shown in Figure~\ref{fig:overview}(b). In this way, the short-term spatio-temporal correlations can be captured in shallower layers while the long-term dynamics can be extracted in deeper layers. Both short-term and long-term spatio-temporal dependencies can be taken into account in this framework.



\subsubsection{Graph Structure Search}
For each Auto-DSTSG block, the most important part is graph structure search operation $\mathcal{F}_{G}(\cdot)$, which can flexibly construct the adjacency matrix of spatio-temporal synchronous graph (STSG). According to~\cite{stsgcn, stfgnn}, there are three constraints to construct STSG:
a) the meta graph on the main diagonal in STSG must be spatial graph (SG) or temporal graph (TG). b) the matrix on the sub diagonal in STSG must not be the zero matrix. c) the complete adjacency matrix of STSG is assumed to be symmetric. 

\begin{figure}[h]
\centering
\vspace{-3mm}
\includegraphics[width=0.46 \textwidth]{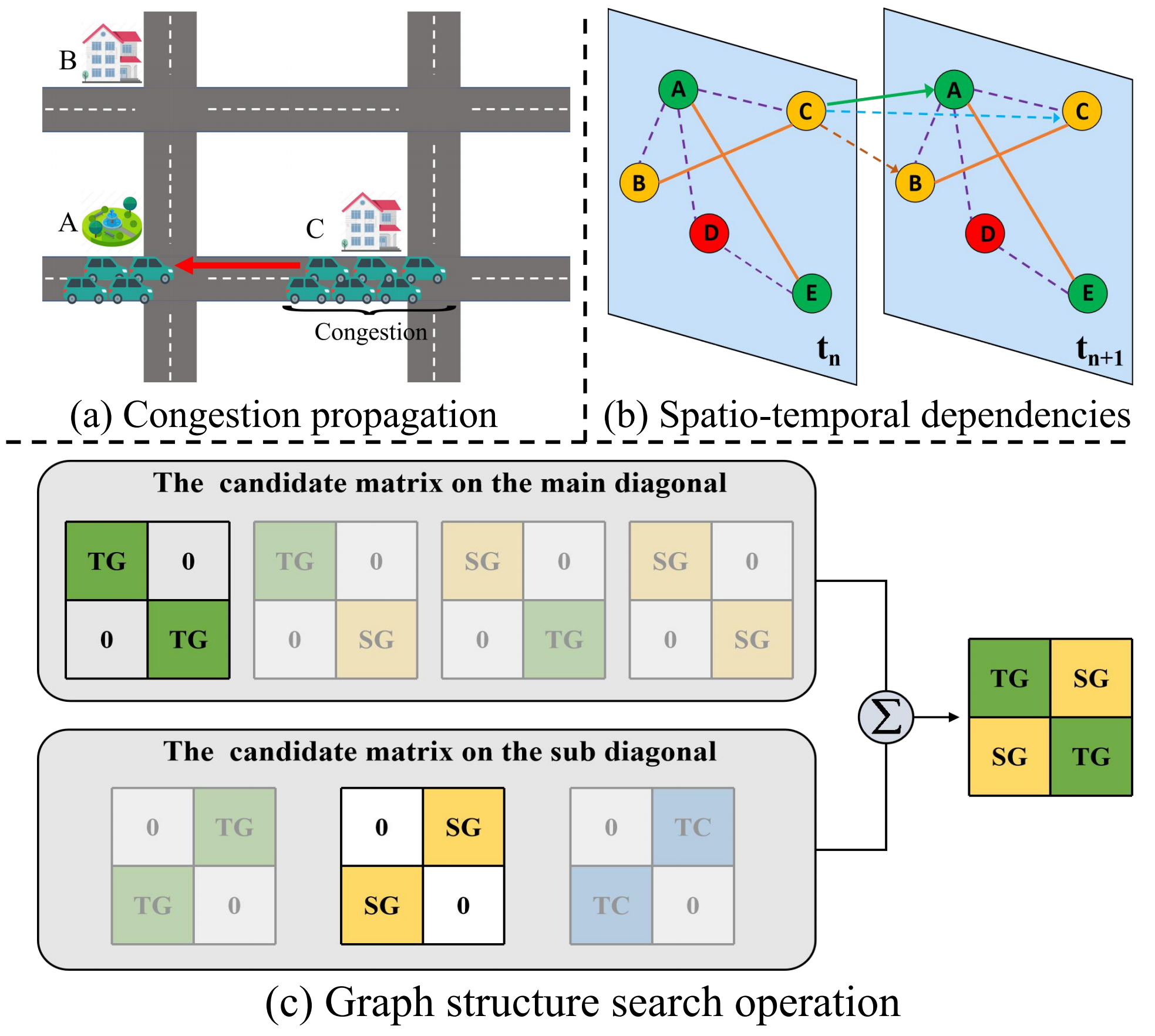}
\caption{Illustration of Graph Structure Search operation $\mathcal{F}_{G}(\cdot)$ with $k=2$ based on a given example. }
\label{fig:gss} 
\vspace{-5mm}
\end{figure}
The meta graph on the main diagonal in STSG can determine the different dependencies in different time steps while the meta graph on the sub diagonal control the correlations between different time steps.
When the size of STSG is fixed as $2N\times2N$, there are four possible options on the main diagonal, they are respectively $[TG,TG]$, $[TG,SG]$, $[SG,TG]$, $[SG,SG]$. There are also three possible options on the sub diagonal, they are respectively $TG$, $SG$, $TC$, where $TC$ denotes the identity matrix to describe self-connectivity. We divide the complete adjacency matrix into two groups by the candidate sub-matrix on the main diagonal and on the sub diagonal. 
It is worth mentioning that this grouping method can also be extended to scenes where $k$ is larger. The search space of these two groups can also be determined by the possible options we discuss above. 

In Figure~\ref{fig:gss}, we take an example to illustrate the selection process of graph structure search operation. Suppose in a traffic scenario during rush hours as shown in Figure~\ref{fig:gss}(a), the pattern similarities between the districts with the same function could be more significant (e.g., district $B$ and $C$). Thus, the temporal graph can better characterize the dependencies in each time step, which can be selected as the meta graph on the main diagonal in STSG. Assume that a congestion event occurs in $C$, the congestion will propagate from $C$ to the spatial adjacent node $A$ over time, as the red arrow shown in Figure~\ref{fig:gss}(a). Thus, the spatial graph can better characterize the dependencies across time steps, which can be selected as the meta graph on the sub diagonal. As shown in Figure~\ref{fig:gss}(b), the nodes with similar patterns are marked by the same color, the significant dependencies are depicted by solid lines and other insignificant dependencies are depicted by dotted lines. Finally, we sum the selected options on main diagonal and sub diagonal to obtain the complete STSG, as shown in  Figure~\ref{fig:gss}(c).

In this case, we design the two-group graph structure search method inspired by DARTS framework~\cite{liu2018darts}. Although adjacency matrices of STSGs are not neural architectures, they can determine the mode of message passing in graph convolution networks. Thus, the structure of the spatio-temporal synchronous adjacency matrices can be seen as a special architecture in our model. 
Similar to the mixed operation in classical DARTS framework, the mixed adjacency matrix generated in searching phase is formulated as follow:
\begin{align}
	&\mathcal{M}_{1} = \sum_{m_{1} \in \mathbf{M_{1}}} \frac{\exp(\alpha_{m_1})}{\sum_{m_{1}' \in \mathbf{M_{1}}} \exp(\alpha_{m_{1}'})} m_{1} \\
	&\mathcal{M}_{2} = \sum_{m_{2} \in \mathbf{M_{2}}} \frac{\exp(\alpha_{m_{2}})}{\sum_{m_{2}' \in \mathbf{M_{2}}} \exp(\alpha_{m_{2}'})}m_{2},\\
	&\mathcal{M}_{A} = \mathcal{M}_{1} + \mathcal{M}_{2},
	\label{eq:soft}
\end{align}
where $m_{1} \in \mathbf{R}^{2N\times 2N}$ and $m_{2} \in \mathbf{R}^{2N\times 2N}$ are respectively the candidate matrix of the case of main diagonal and the sub-diagonal, $\mathbf{M_{1}} = \{M_1^{(1)}, M_1^{(2)}, \cdots \}$ and $\mathbf{M_{2}} = \{M_2^{(1)}, M_2^{(2)}, \cdots \}$ are the set of pre-defined candidate matrices of the two groups. $\mathcal{M}_{A}$ is the complete mixed adjacency matrix as the final output from GSS operation in searching phase. $\alpha_{m_1}$ and $\alpha_{m_2}$ are respectively the learnable parameters to weight the candidate matrix $m_{1}$ and $m_{2}$. 

For each group, the best sub-graph structure is determined by the lowest validation loss in candidate set.
And during the training phase, we replace each mixed operation $\mathcal{M}_{1}$ and $\mathcal{M}_{2}$ with the highest confidence to get the final graph structure $A_{ST}$, which can be formulated as follow:
\begin{align}
	A_{ST} = (\mathrm{Argmax}_{m_{1} \in \mathbf{M_{1}}} \enskip \alpha_{m_1}) + (\mathrm{Argmax}_{m_{2} \in \mathbf{M_{2}}} \enskip \alpha_{m_2}).
	\label{eq:final architecture}
\end{align}

\subsubsection{Mixed-Hop Graph Convolution}
With the mixed adjacency matrix $\mathcal{M}_{A}$ during searching phase or the final adjacency matrix $A_{ST}$ during training phase, we also adopt the mixed-hop mechanism to capture complex spatio-temporal information from different hops. 
In addition, gated linear unit (GLU) is used as the mapping function for each hop. The graph convolution operation for each hop can be formulated as: 
\begin{equation}
    \mathcal{H}_{i} = GLU_{i}((\mathcal{C}(\mathbb{A}^{i}\cdot X))),
\end{equation}
where $\mathbb{A}^{i}\in R^{2N\times 2N}$ denotes the $i_{th}$ hop adjacency matrix during searching phase or training phase, $X\in R^{2N\times D}$ denotes the input data and $\mathcal{H}_{i}\in R^{N\times D}$ is the output of the $i_{th}$ hop graph. $GLU_{i}(\cdot)$ is the individual operation for the $i_{th}$ hop graph convolution, which is formulated as:
\begin{equation}
GLU_{i}(X) = (X_c\cdot W_1 + b_1) \odot \sigma(X_c\cdot W_2 + b_2), 
\end{equation}
where $W_1, W_2 \in \mathbb{R}^{D\times D'}$, $b_1, b_2 \in \mathbb{R}^{D'}$ are weights and bias of GLU, $\odot$ represents element-wise product, $\sigma$ denotes the Sigmoid function and $X_c \in R^{N\times D}$ is the output from the cropping operation. 
Finally, we adopt the max-pooling approach to aggregate the graph information from different hops as the output of the Auto-DSTSG block, which is defined as:
\begin{align}
	\mathcal{H}_{\mathbf{g}} =  MaxPooling([ \mathcal{H}_1, \cdots, \mathcal{H}_i]) \in R^{N\times D}.
	\label{eq:final fusion}
\end{align}
The spatio-temporal input data is treated by multiple Auto-DSTS blocks independently in parallel in each layer. The output from the Auto-DSTSGC module in each layer is formulated as:
\begin{align}
    X_{\mathbf{g}} = [\mathcal{H}_{\mathbf{g}}^{0},\dots, \mathcal{H}_{\mathbf{g}}^{T-d-1}] \in R^{N\times (T-d)\times D},
	\label{eq:output_layer}
\end{align}
where $T$ denotes the time steps of input data and $\mathcal{H}_{\mathbf{g}}^{i}$ denotes the output from $i_{th}$ Auto-DSTSG block.

\subsection{Dilated Temporal Convolution Module}

The weight sharing mechanism in temporal convolution is conducive to learning global temporal dependencies for individual nodes~\cite{stfgnn,cnnshare}. Although Auto-DSTSG module can flexibly capture the complex spatio-temporal correlations in a unified framework, its weight non-sharing mechanism for different blocks is more conducive to capturing diverse local spatio-temporal dependencies for different time periods rather than global dependencies for each node.

The dilated temporal convolution operation is defined as:
\begin{equation}
    \mathbf{x}\star \mathbf{f}(t) = \sum_{s=0}^{k-1} \mathbf{f}(\Theta)\mathbf{x}(t-d\times s),
\end{equation}
where $\mathbf{x}\in \mathbf{R}^T$ denotes the given 1D sequence input, $\mathbf{f} \in \mathbf{R}^K$ denotes a temporal convolutional filter at step $t$, $\Theta$ denotes the learnable weights of the filter and $d$ denotes the dilation factor.  Similar to~\cite{yu2016multi}, we obtain the larger receptive field by expanding the dilation factor in temporal convolution when the layer goes deeper. 
To keep the consistency of the receptive field in each layer, the kernel size $k$ of temporal convolution is fixed as 2 with a sequence of dilation factors $[1,2,4,4]$.

To better control the information flow and reserve the useful information, we adopt the gating mechanism in this case. Similar to \cite{dauphin2017language}, a simple gated temporal convolution network only contains an output gate, which is expressed as:  
\begin{equation}
    X_{\mathbf{f}} = tanh(\mathbf{\Theta_1}\star X+\mathbf{b_1})\odot\sigma(\mathbf{\Theta_2}\star X +\mathbf{b_2}),
\end{equation}
where $X\in R^{N\times T\times D}$ is the given input, $X_{\mathbf{f}}\in R^{N\times (T-k-d+2)\times D'}$ is the output, 
$\mathbf{\Theta_1}$, $\mathbf{\Theta_2}$, $\mathbf{b_1}$ and $\mathbf{b_2}$ are model parameters, $\odot$ is the element-wise product, $tanh(\cdot)$ is the tanh activation function of the outputs, and $\sigma(\cdot)$ is the sigmoid function which controls the ratio of information flow put forward to the next layer.

To enhance the global temporal correlations for individual nodes, we aggregate the output of the dilated temporal convolution module and the output of Auto-DSTSG module in each layer. The output of each layer in Auto-DSTSGN can be defined as:
\begin{equation}
    X_{\mathbf{out}} = \mathbf{Agg}(X_{\mathbf{g}},\ X_{\mathbf{f}}),
\end{equation}
where $X_{\mathbf{out}}$ denotes the output from each layer, $X_{\mathbf{g}}$ and $X_{\mathbf{f}}$ are respectively the output of Auto-DSTSG module and dilated temporal convolution module. $\mathbf{Agg}(\cdot)$ denotes the aggregation function, which is set as sum function in our model. 

\subsection{Other Components}




\subsubsection{Residual connection}
Residual connection is an effective approach to overcome the problem of vanishing gradient in deep neural networks. As shown in Figure~\ref{fig:overview}(a), we employ it in each layer of our model, which is defined as:
\begin{equation}
    X_{l} = X + FC(\mathcal{F}(X)),
\end{equation}
where $X$ denotes the original input of each layer, $\mathcal{F}(\cdot)$ denotes the neural network mapping in each layer, $FC(\cdot)$ denotes the linear mapping and $X_{l}$ is the output of each layer.

\subsubsection{Skip connection}
Skip connection is adopted to fully exploit different level information and aggregate them to obtain powerful representation. As shown in Figure~\ref{fig:overview}(a), each layer has a skip connection, which is defined as:
\begin{equation}
    X_{s} = \sum_{i=0}^{l} FC_{i}(\mathcal{F}(X)),
\end{equation}
where $X_{s}$ denotes the final representation from aggregation of skip connection, $FC_{i}(\cdot)$ denotes the linear mapping. 

\subsubsection{Input layer and output layer}
At the top of our proposed framework, we use a fully connected layer to map the raw input into high-dimensional tensor, which can enhance the capability of representation learning in deep neural networks.

In output layer, we adopt a series of two-layer fully connected layers that do not share weights to deal with predictions at different time steps, which is expressed as follows:
\begin{equation}
    \hat{y}_{i} = ReLU(X_{s}\cdot W_{i}^{1} + b_{i}^{1})\cdot W_{i}^{2}+ b_{i}^{2},
\end{equation}
where $X_{s}$ denotes the aggregated representation from skip connection, $\hat{y}_{i}$ denotes the predicted result at $i_{th}$ step. In this manner, we can obtain the next $T$ step predictions by concentrating the predictions at each step together, which is expressed as:
\begin{equation}
    \hat{Y} = [\hat{y}_{1}, \hat{y}_{2}, \dots. \hat{y}_{T}].
\end{equation}
Finally, we select L1 loss as the loss function in our model:
\begin{equation}
    L(Y, \hat{Y}) = \vert Y-\hat{Y}\vert.
\end{equation}

\subsection{Searching Algorithm}
In graph structure search operation, all computations are differentiable. Similar to DARTS framework, a bi-level gradient-based optimization algorithm can be employed to update the weight parameters $\theta$ of the network (including the parameters in TCNs, GLUs, FCs and MLP) and the architecture parameters $\omega$ (including the scores of candidate matrices) alternately. As shown in Algorithm 1, the weight parameters (Line 4-5) and architecture parameters (Line 6-7) are alternately updated based on the training and validation sets, until the stopping criteria is met. Then, the structures of STSGs can be obtained by selecting the candidate operations with the highest operation scores.

\begin{algorithm}[htb]  
  \caption{Optimization algorithm of Auto-DSTSGN.} 
  \label{alg:Framwork}  
  \begin{algorithmic}[1]
    \Require  
      Traffic data from road networks:$[V_1,\dots, V_{N}]$, adjacency matrix of spatial graph $A_s$ and temporal graph $A_t$
    \Ensure  
      The learned STSGs in each Auto-DSTSG module and the predictions in next 12 steps;  
    \State Build $\mathbb D_{train}$, $\mathbb D_{valid}$ from $[V_1,\dots, V_{N_{e}}]$, $A_s$ and $A_t$;  
    \State Initialize the graph structure parameters $\omega$ and the weight parameters $\theta$;  
    \State \textbf{do:} 
    \State \quad Sample $\mathbb D_{batch}$ from $\mathbb D_{train}$;  
    \State \quad $\theta \leftarrow \theta - \mu_{\theta} \nabla_{\theta} L(\theta,\omega,D_{batch})$, $\mu_{\theta}$ is learning rate  
    \State \quad Sample $\mathbb D_{batch}$ from $\mathbb D_{valid}$;    
    \State \quad $\omega \leftarrow \omega - \mu_{\theta} \nabla_{\omega} L(\theta, \omega,D_{batch})$, $\mu_{\theta}$ is learning rate;
    \State \textbf{until} stopping criteria is met;
    \State Get the graph structures, and further train the model on $\mathbb D_{train}$
  \end{algorithmic}  
\end{algorithm}

\section{Experiments}

In this section, we conduct extensive experiments on four public datasets to answer the following research questions:
\begin{itemize}
    \item \textbf{RQ1:} How does our proposed Auto-DSTSGN perform compared with the state-of-the-art baselines in traffic prediction?
    \item \textbf{RQ2:} How does our model perform compared with different variants in the ablation study?
    \item \textbf{RQ3:} What kind of graph structures can be searched by GSS mechanism in our model? For different datasets, what are the differences in searching results? 
    \item \textbf{RQ4:} How does the efficiency and GPU occupancy of our model compare with other models?
    \item \textbf{RQ5:}  How do the model parameters (e.g., the max hop of GCN layers) affect the performance of our model?   
\end{itemize}

\subsection{Datasets and Settings}
We evaluate our model on PEMS03, PEMS04, PEMS07 and PEMS08 which are collected from Caltrans Performance Measurement System (PeMS). 
The time granularity of all datasets is set to 5 minutes. 
The spatial graph for each dataset are constructed based on road network topology. 
Z-score normalization is applied to the traffic flow data. Detailed statistics of datasets are shown in Table~\ref{tab:data}.
\begin{table}[h]
\caption{Dataset description and statistics.}
\vspace{-3mm}
	\scalebox{1.0}{
	\begin{tabular}{ccccc}
		\hline
		Datasets  & \#Nodes & \#Edges & \#TimeSteps &\#TimeRange \\ \hline
		PEMS03    & 358     & 547     & 26208     & 9/1/2018 - 11/30/2018       \\
		PEMS04    & 307     & 340     & 16992     & 1/1/2018 - 2/28/2018        \\
		PEMS07    & 883     & 866     & 28224     & 5/1/2017 - 8/31/2017        \\
		PEMS08    & 170     & 295     & 17856     & 7/1/2016 - 8/31/2016       \\ \hline
	\end{tabular}}
	\label{tab:data}
	\vspace{-3mm}
\end{table}

\begin{table*}[!htb]
	\centering
	\caption{Performance comparison of baseline models and Auto-DSTSGN. The sub- optimal results are marked by the asterisk}
 	\vspace{-3mm}
	\scalebox{0.78}{
		\begin{tabular}{cccccccccccc}
			\hline
			\multirow{2}{*}{\shortstack{Datasets}} &  \multirow{2}{*}{\shortstack{Metric}}                            & \multirow{2}{*}{\shortstack{FC-LSTM}} & \multirow{2}{*}{DCRNN} & \multirow{2}{*}{STGCN} & \multirow{2}{*}{\shortstack{ASTGCN(r)}} & \multirow{2}{*}{\shortstack{GWN}} & \multirow{2}{*}{STSGCN} & \multirow{2}{*}{STFGNN} &\multirow{2}{*}{STGODE} & \multirow{2}{*}{AutoSTG} & \multirow{2}{*}{Auto-DSTSGN}\\
			&                                              &                          &                        &                        &                            &                               &                         &                         \\ \hline
			\multicolumn{1}{c|}{\multirow{3}{*}{\shortstack{PEMS03}}} & \multicolumn{1}{c|}{MAE}                      & 21.33 $\pm$ 0.24                    & 18.18 $\pm$ 0.15                  & 17.49 $\pm$ 0.46                  & 17.69 $\pm$ 1.43                      & {19.85 $\pm$ 0.03}                & 17.48 $\pm$ 0.15                   & {16.77 $\pm$ 0.09}  & {16.53 $\pm$ 0.10} & {16.27* $\pm$ 0.27}   &  \textbf{14.59 $\pm$ 0.05}     \\ 
			\multicolumn{1}{c|}{}                        & \multicolumn{1}{c|}{MAPE(\%)}                & 23.33 $\pm$ 1.23                    & 18.91 $\pm$ 0.82                  & 17.15 $\pm$ 0.45                  & 19.40 $\pm$ 2.24                      & {19.31 $\pm$ 0.49}                & 16.78 $\pm$ 0.20                  & {16.30$\pm$ 0.09}   & {16.68$\pm$ 0.05} & {16.10* $\pm$ 0.03}      & \textbf{14.22 $\pm$ 0.16}     \\ 
			\multicolumn{1}{c|}{}                        & \multicolumn{1}{c|}{RMSE}                & 35.11 $\pm$ 0.50                   & 30.31 $\pm$ 0.25                  & 30.12 $\pm$ 0.70                 & 29.66 $\pm$ 1.68                      & {32.94 $\pm$ 0.18}                & 29.21 $\pm$ 0.56                   & {28.34$\pm$ 0.46}    & {27.79 $\pm$ 0.32}   & {27.63* $\pm$ 0.78}       & \textbf{25.17 $\pm$ 0.24}    \\ \hline
			\multicolumn{1}{c|}{\multirow{3}{*}{\shortstack{PEMS04}}} & \multicolumn{1}{c|}{MAE}                & 27.14 $\pm$ 0.20                   & 24.70 $\pm$ 0.22                  & 22.70 $\pm$ 0.64                  & 22.93 $\pm$ 1.29                      & {25.45 $\pm$ 0.03}                & 21.19 $\pm$ 0.10                  & {19.83*$\pm$ 0.06}  & {20.84$\pm$ 0.07}  & {20.38 $\pm$ 0.09}    & \textbf{18.85 $\pm$ 0.08}    \\ 
			\multicolumn{1}{c|}{}                        & \multicolumn{1}{c|}{MAPE(\%)}                & 18.20 $\pm$ 0.40                    & 17.12 $\pm$ 0.37                  & 14.59 $\pm$ 0.21                 & 16.56 $\pm$ 1.36                      & {17.29 $\pm$ 0.24}                & 13.90 $\pm$ 0.05                   & \textbf{13.02$\pm$ 0.05} & 13.76 $\pm$ 0.04  & {14.12 $\pm$ 0.02}   & 13.21* $\pm$ 0.02      \\ 
			\multicolumn{1}{c|}{}                        & \multicolumn{1}{c|}{RMSE}                & 41.59 $\pm$ 0.21                   & 38.12 $\pm$ 0.26                  & 35.55 $\pm$ 0.75                  & 35.22 $\pm$ 1.90                      & {39.70 $\pm$ 0.04}                & 33.65 $\pm$ 0.20                   & {31.88*$\pm$ 0.14} & {32.84 $\pm$ 0.19} & {32.51 $\pm$ 0.12}  & \textbf{30.48 $\pm$ 0.17}        \\ \hline
			\multicolumn{1}{c|}{\multirow{3}{*}{\shortstack{PEMS07}}} & \multicolumn{1}{c|}{MAE}                & 29.98 $\pm$ 0.42                    & 25.30 $\pm$ 0.52                  & 25.38 $\pm$ 0.49                 & 28.05 $\pm$ 2.34                      & {26.85 $\pm$ 0.05}                & 24.26 $\pm$ 0.14                   & {22.07*$\pm$ 0.11}  & {23.02$\pm$ 0.15} & {23.22 $\pm$ 0.33}    & \textbf{20.08 $\pm$ 0.08}     \\ 
			\multicolumn{1}{c|}{}                        & \multicolumn{1}{c|}{MAPE(\%)}                & 13.20 $\pm$ 0.53                    & 11.66 $\pm$ 0.33                  & 11.08 $\pm$ 0.18                 & 13.92 $\pm$ 1.65                     & {12.12 $\pm$ 0.41}                & 10.21 $\pm$ 1.65                   & {9.21*$\pm$ 0.07}  & {10.09$\pm$ 0.09}    & {9.95 $\pm$ 0.01}   & \textbf{8.57 $\pm$ 0.05}    \\ 
			\multicolumn{1}{c|}{}                        & \multicolumn{1}{c|}{RMSE}                & 45.94 $\pm$ 0.57                    & 38.58 $\pm$ 0.70                 & 38.78 $\pm$ 0.58                 & 42.57 $\pm$ 3.31                      & {42.78 $\pm$ 0.07}                & 39.03 $\pm$ 0.27                   & {35.80*$\pm$ 0.18}  & {37.48 $\pm$ 0.39} & {36.47 $\pm$ 0.47}  & \textbf{33.02 $\pm$ 0.12}       \\ \hline
			\multicolumn{1}{c|}{\multirow{3}{*}{\shortstack{PEMS08}}} & \multicolumn{1}{c|}{MAE}                & 22.20 $\pm$ 0.18                   & 17.86 $\pm$ 0.03                  & 18.02 $\pm$ 0.14                  & 18.61 $\pm$ 0.40                      & {19.13 $\pm$ 0.08}                & 17.13 $\pm$ 0.09                   & {16.64$\pm$ 0.09}  & {16.79$\pm$ 0.08}  & {16.37* $\pm$ 0.12}    & \textbf{14.74 $\pm$ 0.04}     \\  
			\multicolumn{1}{c|}{}                        & \multicolumn{1}{c|}{MAPE(\%)}               & 14.20 $\pm$ 0.59                    & 11.45 $\pm$ 0.03                 & 11.40 $\pm$ 0.10                  & 13.08 $\pm$ 1.00                      & {12.68 $\pm$ 0.57}                & 10.96 $\pm$ 0.07                   & {10.60$\pm$ 0.06}   & {10.58$\pm$ 0.04}  & {10.36* $\pm$ 0.03}  & \textbf{9.45 $\pm$ 0.03}    \\ 
			\multicolumn{1}{c|}{}                        & \multicolumn{1}{c|}{RMSE}               & 34.06 $\pm$ 0.32                   & 27.83 $\pm$ 0.05                  & 27.76 $\pm$ 0.20                  & 28.16 $\pm$ 0.48                     & {31.05 $\pm$ 0.07}                & 26.80 $\pm$ 0.18                   & {26.22$\pm$ 0.15}   & {26.01 $\pm$ 0.14}   & {25.46* $\pm$ 0.18}  & \textbf{23.76 $\pm$ 0.05}     \\ \hline
	\end{tabular}}
	\vspace{-3mm}
	\label{tab:Performance}
\end{table*}

\begin{table}[!htb]
\caption{Ablation experiments.}
\vspace{-3mm}
\centering
\scalebox{1.1}{
	\begin{tabular}{clccc}
		\hline
		Dataset                 & Model\&Variants    & MAE            & MAPE(\%)         & RMSE           \\ \hline
		\multirow{6}{*}{PEMS04} & Auto-DSTSGN        & \textbf{18.85}          & \textbf{13.21}          & \textbf{30.48}          \\
		& w/o GCN   & 23.45 & 16.96 & 36.53 \\
		& w/o TCN    & 19.33 & 14.12 & 30.79 \\
		& w/o Dilation  & 19.18 & 13.76 & 30.71 \\
		& w/o GSS  & 19.48 & 14.06 & 30.72 \\ 
		& GGS Random  & 19.68 & 14.31 & 31.26 \\ 
		& w/o DTW  & 19.27 & 13.84 & 31.05\\ \hline
		\multirow{6}{*}{PEMS08} & Auto-DSTSGN        & \textbf{14.74}          & \textbf{9.45}          & \textbf{23.76}          \\
		& w/o GCN   & 18.59 & 12.73 & 28.91 \\
		& w/o TCN    & 15.22 & 10.12 & 24.02 \\
		& w/o Dilation  & 15.10 & 10.16 & 23.95 \\
		& w/o GSS  & 15.31 & 10.15 & 24.21 \\ 
		& GGS Random  & 15.31 & 10.15 & 24.21 \\ 
		& w/o DTW  & 15.07 & 9.97 & 24.10\\ \hline
	\end{tabular}}
	\label{tab:ablation}
	\vspace{-2mm}
\end{table}

Each dataset is chronologically split with 60\% for training, 20\% for validation and 20\% for testing.
We use the historical traffic flow in the last one hour to forecast the future traffic flow in the next one hour. The spatial graph and temporal graph are designed inspired by~\cite{stfgnn}. Our model is implemented by Pytorch 1.5 with NVIDIA TESLA V100 GPU. 
The max hop of graph convolution in each Auto-DSTSG block is set as 2 by default. The dimension of hidden representations is set as 40. 
The optimizer of our model is set as Adam. The batch size is 64 and the learning rate is 0.001. 
Our model is evaluated five times on each dataset. During search process, we utilize the early stopping strategy for graph structure search with tolerance 15 for 60 epochs. During training process, we reinitialize the optimizer and employ early stopping with tolerance 30 for 200 epochs. 

\subsection{Overall Performance (RQ1)}

We compare our model with the following nine state-of-art baselines in recent years:
\begin{itemize}
\item \textbf{FC-LSTM:} This is a variant of Long Short-Term Memory Network, which adopts fully connected hidden units~\cite{lstm}. We set the number of hidden layer as 1 and the hidden units as 64.

\item \textbf{DCRNN:} Diffusion Convolution Recurrent Neural Network, which integrates GCNs into encoder-decoder RNNs~\cite{dcrnn}. The hop of diffusion graph convolution is set as 2 and the hidden dimension is set as 64. 

\item \textbf{STGCN:} Spatio-Temporal Graph Convolution Network, which employs GCNs and TCNs for spatio-temporal learning~\cite{stgcn}. Each spatio-temporal cell in this model contains two TCNs and one GCN. The number of spatio-temporal cell is set as 2 and the hidden dimension is set as 64. 

\item \textbf{ASTGCN:} Attention based Spatial Temporal Graph Convolution Network, which utilizes spatial and temporal attention mechanisms~\cite{astgcn}. Similar to STGCN, there are two spatio-temporal cells in this model and the hidden dimension is set as 64.

\item \textbf{Graph WaveNet (GWN):} Graph WaveNet adopts the GCNs with adaptive adjacency matrix and 1D dilated TCNs for spatio-temporal modeling~\cite{gwn}. Each layer in this model contains a gated TCN and a spatial GCN. The number of stacked layers in this model is set as 8 with the dilation rate [1, 2, 1, 2, 1, 2, 1, 2, 1, 2] and the hidden dimension is set as 64.

\item \textbf{STSGCN:} Spatial-Temporal Synchronous Graph Convolution Network, which utilizes multiple STSG modules for localized spatio-temporal joint dependencies modeling~\cite{stsgcn}. The size of spatial-temporal synchronous graph is set as $3N\times 3N$, the number of STSG layers is set as 3, and the hidden dimension is set as 64 in this model.

\item \textbf{STFGNN:} Spatial-Temporal Fusion Graph Convolution Network, which utilizes spatio-temporal fusion graph convolution and parallel TCNs for learning localized and global spatio-temporal dependencies respecively~\cite{stfgnn}. The size of spatial-temporal fusion graph is set as $4N\times 4N$, the number of layers is set as 3, and the hidden dimension is set as 64 in this model. 

\item \textbf{STGODE:} Spatial-Temporal Graph Ordinary Differential Equation Network, which integrates the tensor-based ordinary differential equation into GCN modules~\cite{stgode}. The number of graph ODE layers is set as 6 and the hidden dimension is set as 64 in this model.

\item \textbf{AutoSTG:} Automated Spatio-Temporal Graph Network, which integrates NAS with spatio-temporal graph learning modules~\cite{autostg}. The number of spatio-temporal graph learning cells is set as 5, the number of mixed operations is set as 6 in each cells, and the the hidden dimension is set as 64 in this model.
\end{itemize}

The evaluation metrics are mean absolute errors (MAE), root mean squared errors (RMSE) and mean absolute percentage errors (MAPE) averaged over five times for one hour ahead prediction. 
For these three metrics, smaller values means better performance and the formula of these three metrics are defined as follows:
\begin{equation}
{\rm RMSE}(\widehat{y}_{i},y_{i}) =\sqrt{\frac{1}{n}\sum_{i=1}^{n}{(y_{i}-\widehat{y}_i)^2}},
\end{equation} 
\begin{equation}
{\rm MAE}(\widehat{y}_{i},y_{i}) =\frac{1}{n}\sum_{i=1}^{n}{|y_{i}-\widehat{y}_i|}, \quad\quad
\end{equation} 
\begin{equation}
{\rm MAPE}(\widehat{y}_{i},y_{i}) =\frac{1}{n}\sum_{i=1}^{n}{\frac{|y_{i}-\widehat{y}_i|}{y_{i}}}.\quad
\end{equation} 
where $\widehat{y}_{i}$ denotes the prediction results and ${y}_{i}$ denotes the ground-truths.

\begin{figure}[h]
\vspace{-2mm}
\centering
\includegraphics[width=0.48 \textwidth]{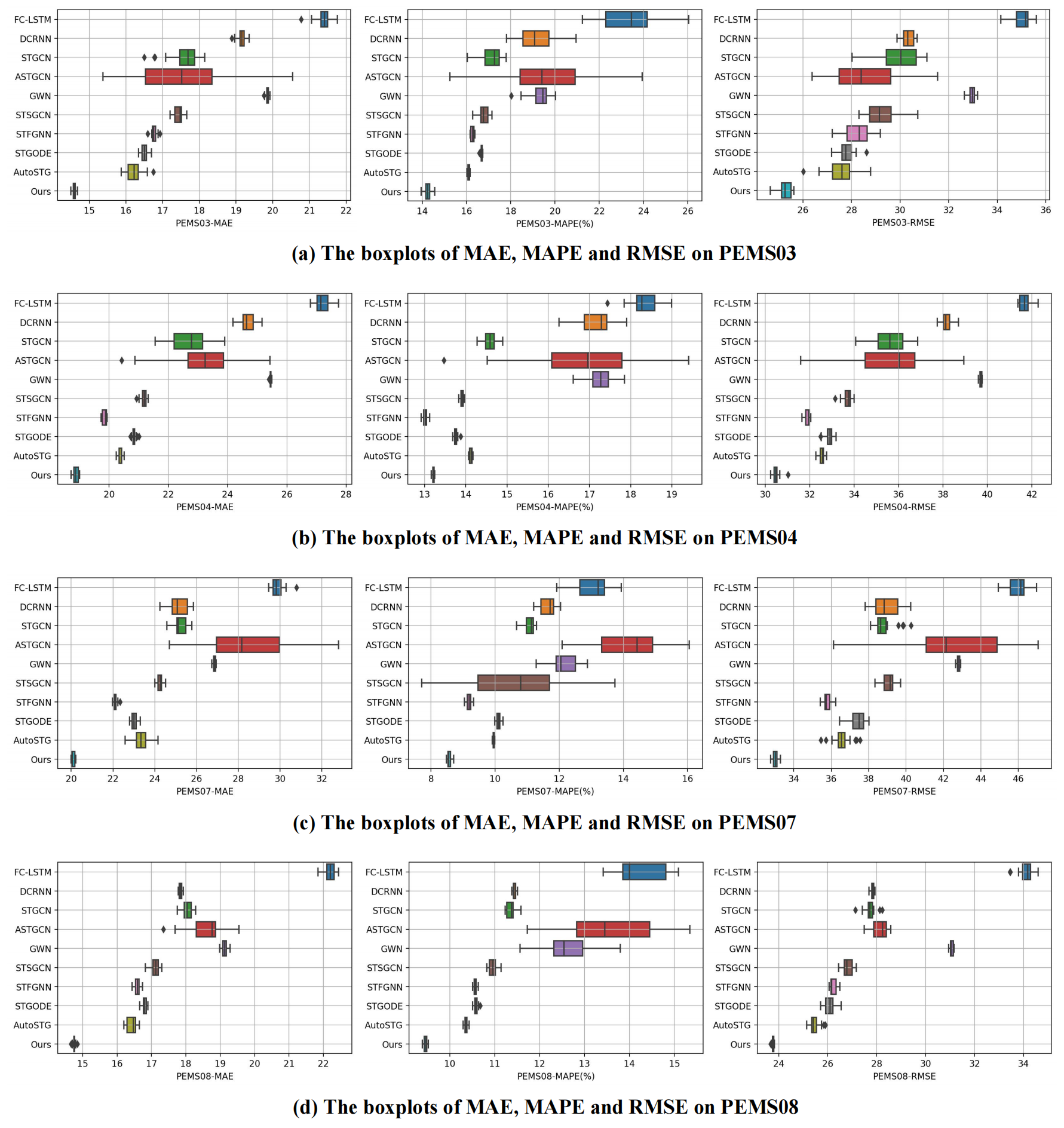}
\caption{Boxplots for the comparison results in Table~\ref{tab:Performance}.}
\label{fig:box} 
\vspace{-4mm}
\end{figure}

From the results in Table~\ref{tab:Performance}, we can observe that our model Auto-DSTSGN consistently outperforms the sub-optimal baselines with 4.9\%$\sim$10.3\% improvements in terms of MAE on the four datasets, which demonstrates the superiority of our proposed method. In order to make the comparison results more intuitive visually, we also provide the box plot of Table~\ref{tab:Performance}, as shown in Figure~\ref{fig:box}. Next, we analyze and compare the strengths of our proposed model with some well-performing baselines.

AutoSTG is the only baseline that integrates the neural architecture search to adjust its architecture corresponding to the data. From the comparison results, AutoSTG significantly outperforms most non-auto state-of-art models such as DCRNN and Graph WaveNet, since it can search the optimal neural architectures for different data scenarios. However, AutoSTG is still weaker than our model. There are two main reasons. The one is that the spatio-temporal synchronous graphs of our model can characterize more complex dynamics than the normal graphs. Another one is that AutoSTG focuses on the neural architecture search but ignores the importance of informative spatio-temporal graph construction.   
Both STFGNN and STSGCN adopt the spatio-temporal synchronous graphs but they can hardly capture the short-term and long-term spatio-temporal correlations flexibly. Our model presents the dilated spatio-temporal synchronous graph convolution framework to balance the short-term and long-term dependencies learning. 
Both STFGNN and STGODE are the baselines that adopt informative adjacency matrices for spatio-temporal dependencies learning. However, these two models design the spatio-temporal synchronous graphs or neural architectures manually and  empirically, thus they are difficult to adapt to different data scenarios. In contrast, our model can construct different spatio-temporal synchronous graphs for different data scenarios based on auto machine learning, which enhances the adaptability and generalizability of our model.

In summary, the dilated spatio-temporal synchronous graph structures in our model can flexibly characterize the short and long-term spatio-temporal dependencies, the auto machine learning mechanism for graph structure search can help our model adapt to different data inputs and achieve the optimal dilated spatio-temporal synchronous graph modeling. These are why our model can outperform other baselines significantly.

\subsection{Ablation Study (RQ2)}
We conduct ablation study on PEMS04 and PEMS08 to evaluate the effectiveness of key components in our model. As shown in Table \ref{tab:ablation}, we compared Auto-DSTSGN with following variants: 1) \emph{w/o GCN}, which removes all the Auto-DSTSG modules from our models 2) \emph{w/o TCN}, which removes all the dilated temporal convolution modules from our model.   
3) \emph{w/o Dilation}, which removes the dilation mechanism in Auto-DSTSG module from our model. 4) \emph{w/o GSS}, which replaces the Auto-DSTSG module with the fixed STFGNN module~\cite{stfgnn}. 5) \emph{GSS Random}, which samples an complete adjacency matrix from search space in graph structure search module during the search process. 6) \emph{w/o DTW}, which replaces the DTW with Pearson coefficient to construct the temporal graphs.

From the experimental results, we can find that Auto-DSTSGN outperforms all the ablation variants. 
Compared with the results of \emph{w/o Dilation}, Auto-DSTSGN improves 1.7\%, 4.0\% in terms of MAE and MAP on PEMS04. Meanwhile, it also improves 2.4\%, 7.0\% in terms of MAE and MAPE on PEMS08, which illustrates the effectiveness of dilation mechanism on learning long-term spatio-temporal dependencies. 
Compared with the results of \emph{w/o GSS}, Auto-DSTSGN improves 3.2\%, 6.1\% in terms of MAE and MAPE on PEMS04. In the meanwhile, it also improves 3.7\%, 6.9\% in terms of MAE and MAPE on PEMS08. Compared with the results of \emph{GSS Random}, Auto-DSTSGN improves 4.2\%, 7.7\% in terms of MAE and MAPE on PEMS04. Meanwhile, it also improves 5.5\%, 12.4\% in terms of MAE and MAPE on PEMS08, which illustrates the effectiveness of Graph structure search module in learning diverse spatio-temporal correlations and adapting to different data. 
There is a significant fall of the performance without graph convolutions and temporal convolutions (\emph{w/o GCN} and \emph{w/o TCN}), demonstrating the effectiveness of these two parts for spatio-temporal representation learning. Moreover, when we replace DTW with Pearson coefficient to construct the temporal graphs (\emph{w/o DTW}), the performance drops on both PEMS04 and PEMS08 datasets. The reason is that Pearson coefficient can only characterize the linear correlations between different time series while DTW can better characterize the non-linear time series similarity.

\subsection{Case Study (RQ3)}

\begin{figure}[h]
\centering
\vspace{-1mm}
\includegraphics[width=0.45 \textwidth]{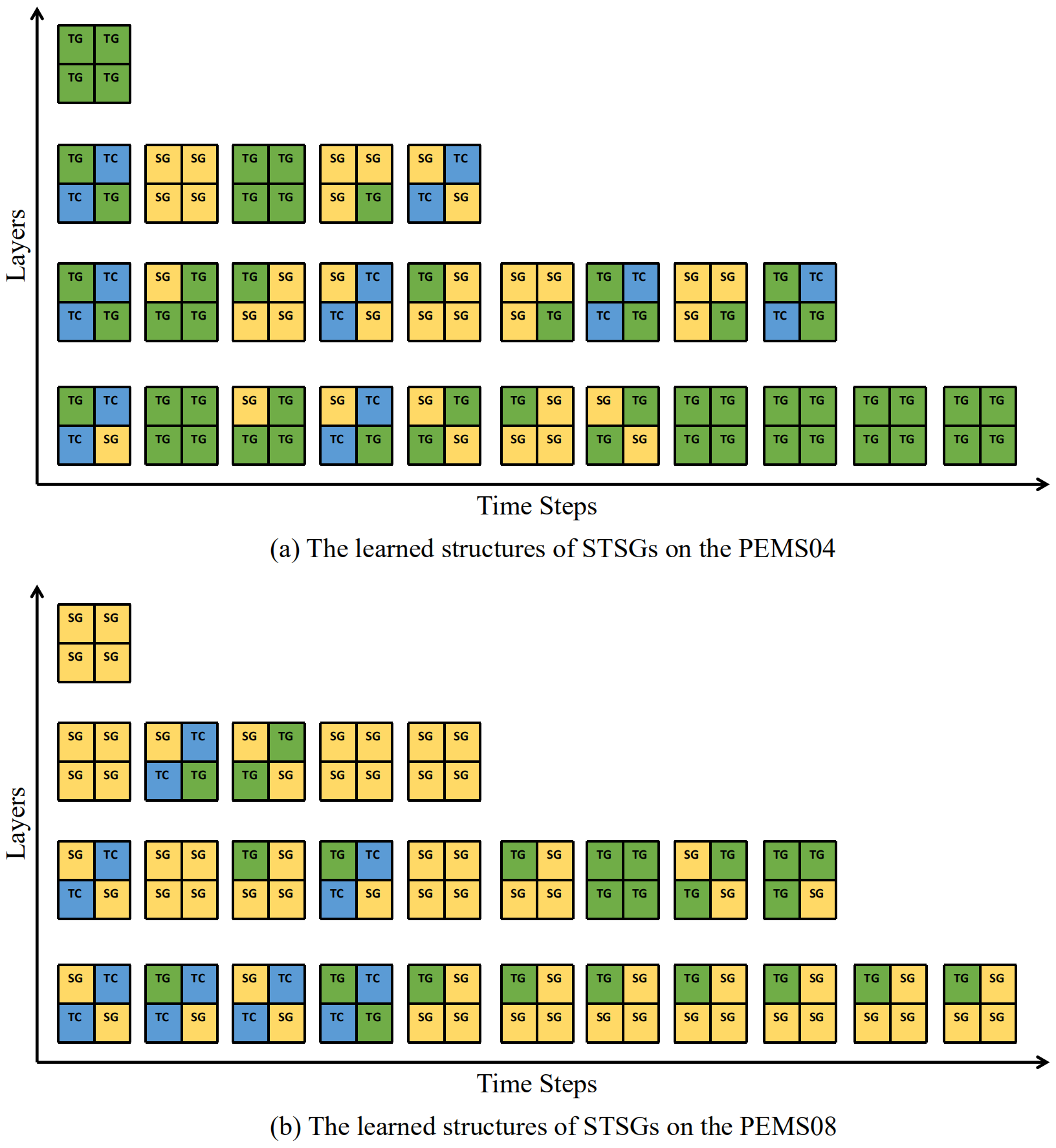}
\caption{The learned structures of STSGs on the PEMS04 and PEMS08.}
\label{fig:stsg} 
\vspace{-3mm}
\end{figure}

We select PEMS04 and PEMS08 to further investigate the relations between the attributes of meta graphs and the optimal structure of STSGs on them. For the two different meta graphs SG and TG, we choose mean degree of them as the most important attribute. The higher mean degree of a graph means stronger correlations among the nodes. Specifically, the higher mean degree of SG represents the stronger spatial correlations while the higher mean degree of TG represents the stronger temporal correlations. 
Thus, empirically we need more graph convolution operations on the related graphs to achieve a larger receptive field for capturing long-range spatio-temporal correlations.
We count the average number of SG and TG in the learned structure of STSGs on two datasets. All the results are shown in Table~\ref{tab:case study}. We observe that the STSGs learned on PEMS04 contains more TGs while the STSGs learned on PEMS08 contains more SGs. This can be explained by the mean degree of TG and SG on two datasets. Since the higher mean degree characterizes stronger correlations between different nodes, our model can obtain stronger capability for message passing in spatio-temporal scale by adjusting the structure of STSGs automatically. In addition, we also visualize the learned structures of STSGs on the PEMS04 and PEMS08 in Figure~\ref{fig:stsg}.
\begin{table}[htb]
\centering
\caption{The attributes of two datasets and the corresponding structures of STSGs.}
\vspace{-3mm}
\scalebox{1.0}{
	\begin{tabular}{clcc}
		\hline
		\textbf{Objects}    & \textbf{Attributes}    & \textbf{PEMS04}    & \textbf{PEMS08} \\ \hline
		\multirow{2}{*}{\textbf{Datasets}} & Mean degree of SG   & 2.21     & 3.22\\
		& Mean degree of TG  & 4.51     & 1.27\\ \hline
		\multirow{2}{*}{\textbf{Structures of STSGs}} & Average \# of SGs  & 35 & 68\\
		& Average \# of TGs   & 47     & 30 \\ \hline
	\end{tabular}}
	\label{tab:case study}
\vspace{-1mm}
\end{table}

\subsection{Efficiency and Occupancy Analysis (RQ4)}
\begin{figure}[h]
\centering
\vspace{-2mm}
\includegraphics[width=0.5 \textwidth]{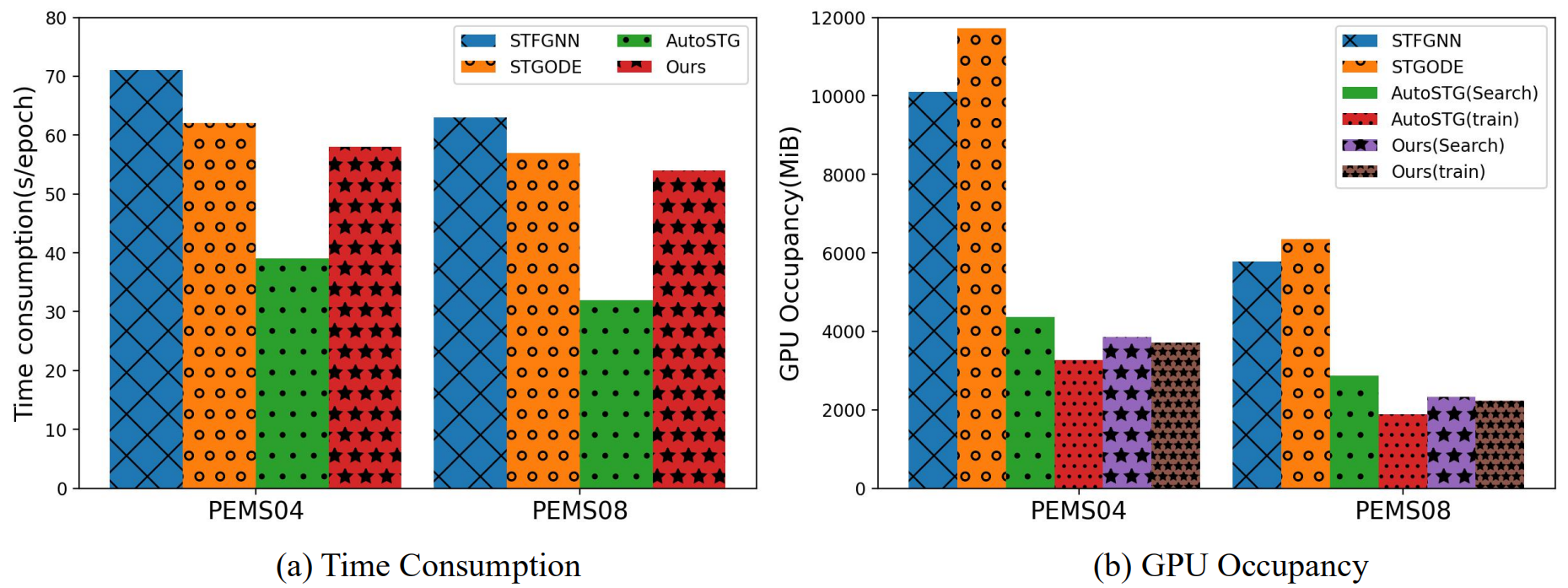}
\caption{Time consumption and GPU occupancy.}
\label{fig:em_study} 
\vspace{-3mm}
\end{figure}

Time consumption and memory occupancy on GPU are two intuitive metrics to reflect the time and space complexity of different models. They are also two important metrics that measure the efficiency and scalability of the model in industrial scenarios. We select three best baselines STFGNN, STGODE and AutoSTG to compare with our model on the two metrics. The results are shown in Figure~\ref{fig:em_study}. From the absolute perspective, the efficiency of our model during training phase is slightly improved compared with STFGNN and STGODE in general and the GPU occupancy of our model is significantly lower than them.  We involve the dilation mechanism into the spatio-temporal synchronous graph modeling, so our model significantly reduces model complexity compared with STFGNN. STGODE employs the multiple neural ODE architectures, which greatly increase the time and space complexity of the model.
From the relative perspective, the GPU occupancy of our model is almost the same during search phase and training phase, but AutoSTG significantly costs more GPU occupancy in the searching phase. This indicates that graph structure search is more lightweight and efficient than neural architecture search. This is because graph structure search only involves simple matrix calculations whereas neural architecture search involves complex computations of neural network structures with learnable parameters. 

\subsection{Parameters Study (RQ5)}
\begin{figure}[h]
\vspace{-3mm}
\centering
\includegraphics[width=0.5 \textwidth]{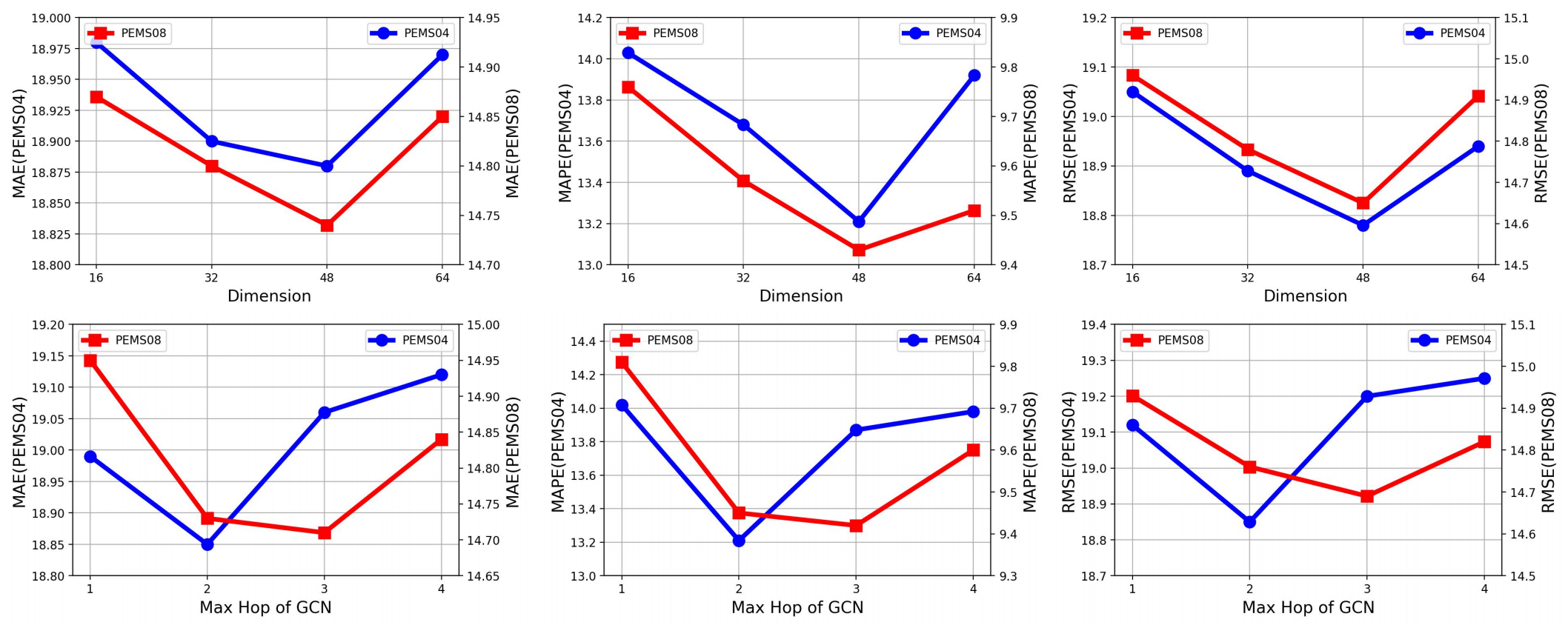}
\caption{Studies on hyper-parameters.}
\label{fig:param} 
\vspace{-3mm}
\end{figure}

To further investigate the effectiveness of our model, we conduct parameter study on PEMS04 and PEMS08, including the dimension of hidden representations $D$ and the max hop of graph convolution $H$ in each block. The experimental results are shown in Figure~\ref{fig:param}. We can find that MAE, MAPE and RMSE on two datasets are the optimal when $D$ is equal to 48. When $D$ is too small, the learning capability of our model become worse, resulting in poor prediction performance. When $D$ is too large, the three metrics on both PEMS04 and PEMS08 become worse. This is because too large hidden dimension cause the over-fitting. For parameter $H$, we can observe that MAE, MAPE and RMSE on PEMS04 achieve the optimal results when $H$ is equal to 2. On PEMS08, MAE, MAPE and RMSE obtain the best values when $H$ is equal to 3. 
This implies that aggregating the neighbor information of the appropriate order in the traffic network can better learn the spatial dependencies, and an excessively large number of hops will lead to the over-smoothing phenomenon of the GCNs.

\section{conclusion}
We propose a novel automated dilated spatio-temporal synchronous graph convolution framework to capture complex spatio-temporal dependencies for traffic prediction. Our model can not only capture complex long-term and short-term spatio-temporal dependencies, but also more flexibly characterize spatio-temporal relations of different time steps and even different scenarios through graph structure search. Extensive experiments on four real-world datasets demonstrate the superiority of our model in prediction accuracy compared with other state-of-art baselines. 
In addition, under the premise of ensuring accuracy, our model also takes into account both efficiency and GPU occupancy, which provides a solid foundation for the deployment of the model in industrial scenarios.
In this paper, we give a first attempt to adopt automatic machine learning approach in graph structure search for diverse spatio-temporal relations. In future work, we will extend this method to a more generalized spatio-temporal prediction scenario.




%





\ifCLASSOPTIONcaptionsoff
  \newpage
\fi





\bibliographystyle{IEEEtran}
\bibliography{IEEEabrv,Bibliography}

\vfill


\end{document}